%% file: vic-arXiv.tex
\documentclass[10pt,twocolumn,letterpaper]{article}

\usepackage{iccv}
\usepackage{times}
\usepackage{epsfig}
\usepackage{graphicx}
\usepackage{amsmath}
\usepackage{amssymb}
\usepackage{arydshln} 
\usepackage{booktabs} 

\usepackage{multirow}
\usepackage{adjustbox}
\usepackage{enumitem} 

\usepackage[pagebackref=true,breaklinks=true,letterpaper=true,colorlinks,bookmarks=false]{hyperref}
\iccvfinalcopy

\def\iccvPaperID{1873} % *** Enter the ICCV Paper ID here

\ificcvfinal\fi

\begin{document}

\newcommand{\argmaxsmall}{\operatornamewithlimits{argmax}}
\newcommand{\argmax}[1]{\underset{#1}{\operatorname{arg}\,\operatorname{max}}\;}
\renewcommand{\paragraph}[1]{\noindent \textbf{#1}}
\newcommand\paragraphV{\vspace{0.7mm}\paragraph}
\newcommand{\new}[1]{\textcolor{magenta}{#1}}

\title{Action Tubelet Detector for Spatio-Temporal Action Localization}

\author{Vicky~Kalogeiton\textsuperscript{1,2}
\and
Philippe~Weinzaepfel\textsuperscript{3}
\and
Vittorio~Ferrari\textsuperscript{2}
\and 
Cordelia Schmid\textsuperscript{1}
}

\maketitle
\thispagestyle{empty}

\footnotetext[1]{Univ.~Grenoble Alpes, Inria, CNRS, Grenoble INP, LJK, 38000
Grenoble, France.}
\footnotetext[2]{University of Edinburgh}
\footnotetext[3]{Naver Labs Europe}
%%%%%%%%% ABSTRACT
\input{abstract}

%%%%%%%%% BODY TEXT
\input{intro}
\input{related}
\input{method}
\input{experiments}

\input{conclusions}
%\vspace{3mm}

\vspace{0.5mm}
\noindent {\bf Acknowledgments.}
%{\small
This work was supported in part by the ERC grants ALLEGRO and VisCul, the MSR-Inria joint project, a Google research award, a Facebook gift, an Intel gift and an Amazon research award. We gratefully acknowledge the support of NVIDIA with the donation of GPUs.
%}

{\small
\bibliographystyle{ieee}
\bibliography{shortstrings,vic-iccv17-final}
}

\end{document}

%% file: abstract.tex
\begin{abstract}

Current state-of-the-art approaches for spatio-temporal action localization rely on detections at the frame level that are then linked or tracked across time.  In this paper, we leverage the temporal continuity of videos instead of operating at the frame level. We propose the {\bf AC}tion {\bf T}ubelet detector (ACT-detector) that takes as input a sequence of frames and outputs \textit{tubelets}, \ie, sequences of bounding boxes with associated scores. The same way state-of-the-art object detectors rely on anchor boxes, our ACT-detector is based on anchor cuboids. We build upon the SSD framework~\cite{liu16eccv}. Convolutional features are extracted for each frame, while scores and regressions are based on the temporal stacking of these features, thus exploiting information from a sequence. Our experimental results show that leveraging sequences of frames significantly improves detection performance over using individual frames. The gain of our tubelet detector can be explained by both more accurate scores and more precise localization. Our ACT-detector outperforms the state-of-the-art methods for frame-mAP and video-mAP on the J-HMDB~\cite{jhmdb} and UCF-101~\cite{ucf101} datasets, in particular at high overlap thresholds.  

\end{abstract}

%% file: intro.tex
\vspace{-3mm}
\section{Introduction}
\label{sec:intro}

\begin{figure}[t]
\centering
\includegraphics[width=\linewidth]{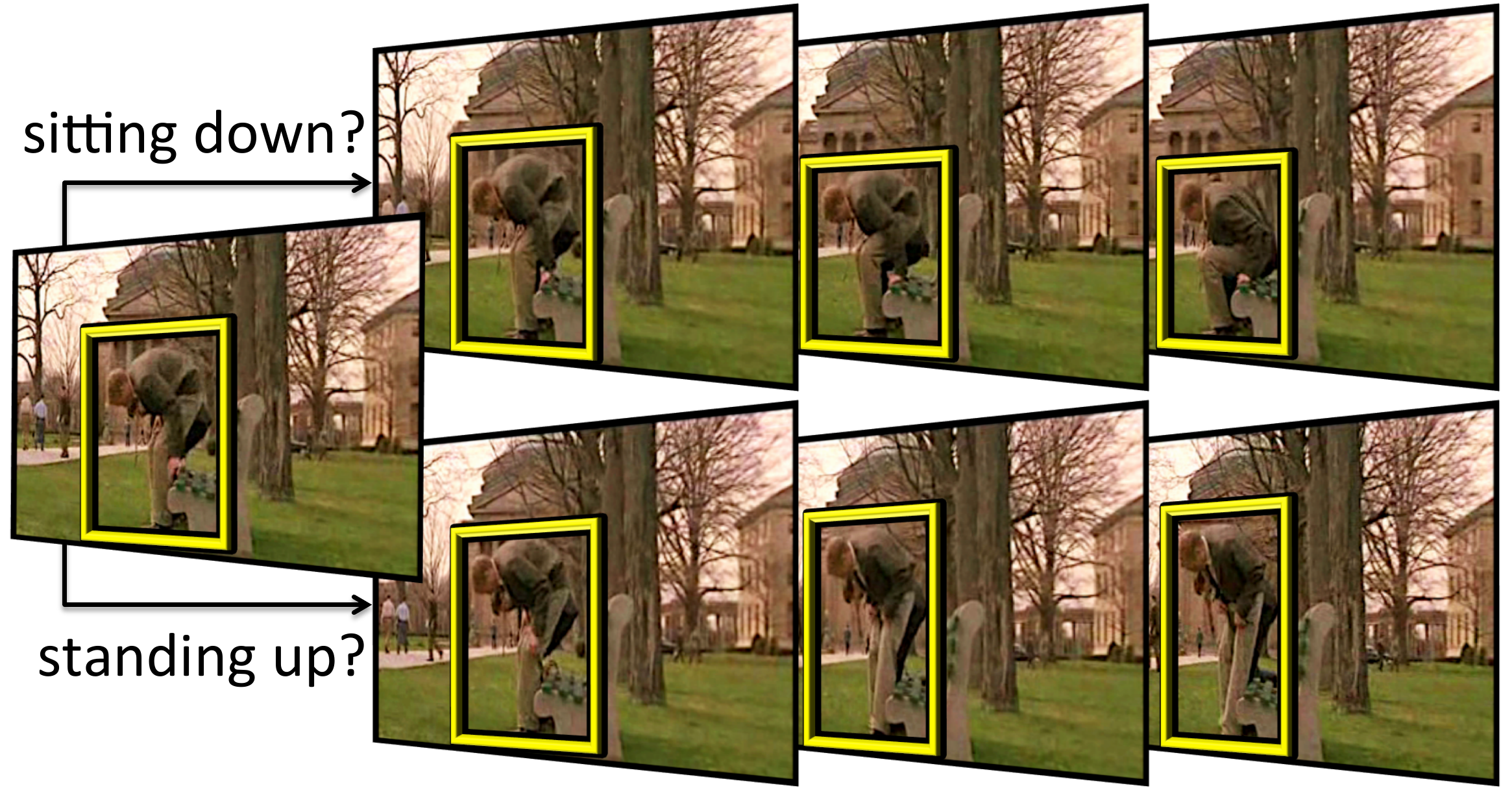}
\vspace{-5mm}
\caption{Understanding an action from a single frame can be ambiguous, \eg \textit{sitting down} or \textit{standing up}; the action becomes clear when looking at a sequence of frames.}
\label{fig:splash}
\vspace{-6mm}
\end{figure}

Action localization is one of the key elements to video understanding. It has been an active research topic for the past years due to various applications, \eg video surveillance~\cite{hu2004survey,oh2011large} or video captioning~\cite{venugopalan2015sequence,yao15iccv}. Action localization focuses both on classifying the actions present in a video and on localizing them in space and time. Action localization task faces significant challenges, \eg intra-class variability, cluttered background, low quality video data, occlusion, changes in viewpoint. Recently, Convolutional Neural Networks (CNNs) have proven well adapted for action localization, as they provide robust representations of video frames. Indeed, most state-of-the-art action localization approaches~\cite{Gkioxari15cvpr,Peng16eccv,Suman16bmvc,singh16arxiv,Weinzaepfel15iccv} are based on CNN object detectors~\cite{liu16eccv,ren15nips} that detect human actions at the frame level. Then, they either link frame-level detections or track them over time to create spatio-temporal tubes. Although these action localization methods have achieved remarkable results~\cite{Peng16eccv,Suman16bmvc}, they do not exploit the temporal continuity of videos as they treat the video frames as a set of independent images on which a detector is applied independently. Processing frames individually is not optimal, as distinguishing actions from a single frame can be ambiguous, \eg \textit{person sitting down} or \textit{standing up} (Figure~\ref{fig:splash}).  

\begin{figure*}[t!]
\centerline{
\includegraphics[width=\linewidth]{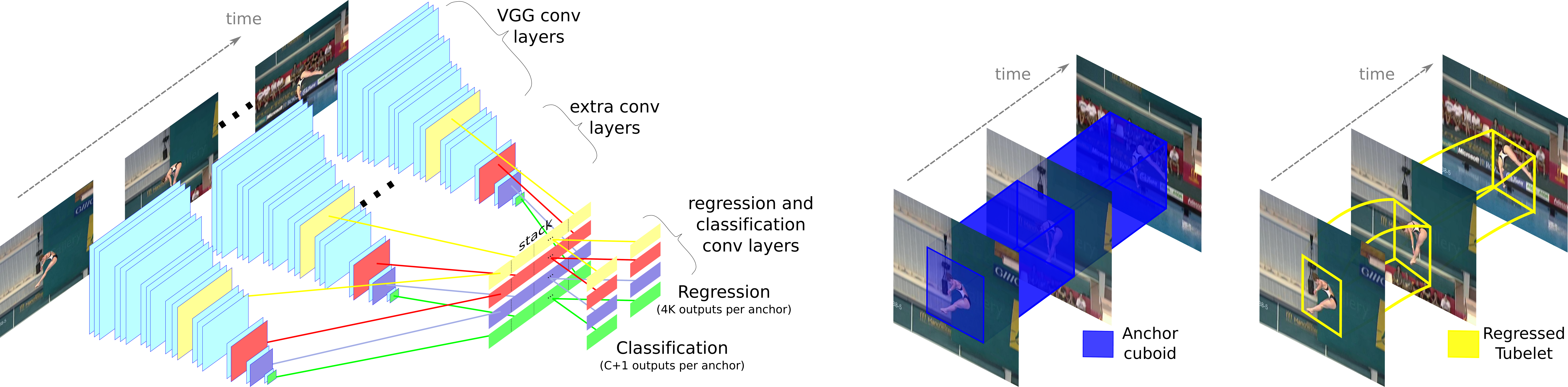}}
\vspace{-1mm}
\caption{Overview of our ACT-detector. Given a sequence of frames, we extract convolutional features with weights shared between frames. We stack the features from subsequent frames to predict scores and regress coordinates for the anchor cuboids (middle figure, blue color). Depending on the size of the anchors, the features come from different convolutional layers (left figure, color coded: yellow, red, purple, green). As output, we obtain tubelets (right figure, yellow color).}
\label{fig:main}
\vspace{-5mm}
\end{figure*}

In this paper, we propose to surpass this limitation and treat a video as a sequence of frames.  
Modern object detectors for images, such as Faster R-CNN~\cite{ren15nips} and Single Shot MultiBox Detector (SSD)~\cite{liu16eccv}, proceed by classifying and regressing a set of anchor boxes to the ground-truth bounding box of the object. 
%For instance, this is the case in the Faster R-CNN detector~\cite{ren15nips} and in the Single Shot MultiBox Detector (SSD) detector~\cite{liu16eccv}. 
In this paper, we introduce a spatio-temporal tubelet extension of this design. 
Our Action Tubelet detector (ACT-detector) takes as input a short sequence of a fixed number of frames and outputs \textit{tubelets}, \ie, sequences of bounding boxes over time (Figure~\ref{fig:main}). Our method considers densely sampled anchors of cuboid shape with various sizes and aspect ratios. 
At test time, we generate for each anchor cuboid a score for a given action and regressed coordinates transforming it into a tubelet. 
Importantly, the score and regression are based on convolutional feature maps from all frames in the sequence. 
While the anchor cuboids have fixed spatial extent across time, the tubelets change size, location and aspect ratio over time, following the actors. Here we build upon the SSD framework, but the proposed tubelet extension is applicable to other detectors based on anchor boxes, such as Faster R-CNN.

Our experiments show that taking as input a sequence of frames improves: (a)~action scoring, because the ambiguity between different actions reduces and (b)~localization accuracy, because frames in a cuboid are regressed jointly and hence, they share information about the location of the actor in neighboring frames, see Figure~\ref{fig:splash}. 
Our ACT-detector obtains state-of-the-art frame-mAP and video-mAP performance on the J-HMDB~\cite{jhmdb} and UCF-101~\cite{ucf101} %action localization
datasets, in particular at high overlap thresholds. %while being faster than previous approaches. 

\noindent In summary, we make the following contributions:

%\noindent $\bullet$ We introduce the ACT-detector, an action tubelet detector for action localization that proceeds by scoring and regressing anchor cuboids.
\noindent $\bullet$ We introduce the ACT-detector, an action tubelet detector that proceeds by scoring and regressing anchor cuboids.

\noindent $\bullet$ We demonstrate that anchor cuboids can handle moving actors for sequences up to around 10 frames. 

\noindent $\bullet$ We provide an extensive analysis demonstrating the clear benefit of leveraging sequences of frames instead of operating at the frame level. 

%After reviewing the related work (Section~\ref{sec:related}), we describe our ACT-detector (Section~\ref{sec:method}). In Section~\ref{sec:experiments}, our experiments demonstrate the benefit of tubelets.

The code of our ACT-detector is available at http://thoth.inrialpes.fr/src/ACTdetector.

%% file: related.tex
%\vspace{-1mm}
\section{Related work}
\label{sec:related}

Almost all recent works~\cite{Peng16eccv,Suman16bmvc,singh16arxiv,Weinzaepfel15iccv} for action localization build on CNN object detectors~\cite{liu16eccv,ren15nips}. In the following, we review recent CNN object detectors and then state-of-the-art action localization approaches.  

\paragraphV{Object detection with CNNs.} 
Recent state-of-the-art object detectors~\cite{girshick14cvpr,liu16eccv,redmon16cvpr,ren15nips} are based on CNNs. R-CNN~\cite{girshick14cvpr} casts the object detection task as a region-proposal classification problem. Faster R-CNN~\cite{ren15nips} extends this approach by generating bounding box proposals with a fully-convolutional Region Proposal Network (RPN). RPN considers a set of densely sampled anchor boxes, that are scored and regressed. Moreover, it shares convolutional features with  proposal classification and regression branches. These branches operate on fixed-size features obtained using a Region-of-Interest (RoI) pooling layer. In a similar spirit, YOLO~\cite{redmon16cvpr} and SSD~\cite{liu16eccv} also use a set of anchor boxes, which are directly classified and regressed without a RoI pooling layer. In YOLO, all scores and regressions are computed from the last convolutional feature maps, whereas SSD adapts the features to the size of the boxes. 
%Features for predicting small-sized boxes come from early layers, and features for big boxes come from the latter layers, which have larger receptive fields.  All these object detectors rely on a set of anchor boxes. In our work, we extend them to anchor cuboids leading to significant improvement for action localization in videos. 
Features for predicting small-sized boxes come from early layers, and features for big boxes come from the latter layers, with larger receptive fields.  All these object detectors rely on anchor boxes. In our work, we extend them to anchor cuboids leading to significant improvement for action localization.

\paragraphV{Action localization.} 
Initial approaches for spatio-temporal action localization are extensions of the sliding window scheme~\cite{msr2,Laptev07iccv}, requiring strong assumptions such as a cuboid shape, \ie, a fixed spatial extent of the actor across frames.  Other methods extend object proposals to videos. Hundreds of action proposals are extracted per video given low-level cues, such as super-voxels~\cite{jain14cvpr,oneata14eccv} or dense trajectories~\cite{chen15iccv,gemert15bmvc,marian15iccv}. They then cast action localization as a proposal classification problem.   
More recently, some approaches~\cite{li16arXiv,wang16cvpr,yu15cvpr} rely on an actionness measure~\cite{chen14cvpr}, \ie, a pixel-wise probability of containing any action. To estimate actionness, they use low-level cues such as optical flow~\cite{yu15cvpr}, CNNs with a two-stream fully-convolutional
architecture~\cite{wang16cvpr} or recurrent neural networks~\cite{wang16cvpr}. They extract action tubes either by thresholding~\cite{li16arXiv} the actionness score or by using a maximum set coverage formulation~\cite{yu15cvpr}. This, however, outputs only a rough localization of the action as it is based on noisy pixel-level maps.  

Most recent approaches rely on object detectors trained to discriminate human action classes at the frame level. Gkioxari and Malik~\cite{Gkioxari15cvpr} extend the R-CNN framework to a two-stream variant~\cite{Simonyan2014nips}, processing RGB and flow data separately. The resulting per-frame detections are then linked using dynamic programming with a cost function based on detection scores of the boxes and overlap between detections of consecutive frames. Weinzaepfel \etal~\cite{Weinzaepfel15iccv} replace the linking algorithm by a tracking-by-detection method. More recently, two-stream Faster R-CNN was introduced by~\cite{Peng16eccv,Suman16bmvc}. Saha \etal~\cite{Suman16bmvc} fuse the scores of both streams based on overlap between the appearance and the motion RPNs. Peng and Schmid~\cite{Peng16eccv} combine proposals extracted from the two streams and then classify and regress them with fused RGB and multi-frame optical flow features. They also use multiple regions inside each action proposal and then link the detections across a video based on spatial overlap and classification score. Singh \etal~\cite{singh16arxiv} perform action localization in real-time using (a)~the efficient SSD detector, (b)~a fast method~\cite{kroeger16eccv} to estimate the optical flow for the motion stream, and (c)~an online linking algorithm. All these approaches rely on detections \textit{at the frame level}. In contrast, we build our ACT-detector by taking as input sequences of frames and demonstrate improved action scores and location accuracy
over frame-level detections.

%% file: method.tex
%\vspace{-1mm}
\section{ACtion Tubelet (ACT) detector}
\label{sec:method}

We introduce the {\bf AC}tion {\bf T}ubelet detector (ACT-detector),  an action tubelet approach for action localization in videos. The ACT-detector takes as input a sequence of $K$ frames $f_1,...,f_K$ and outputs a list of spatio-temporal detections, each one being a \textit{tubelet}, \ie, a sequence of bounding boxes, with one confidence score per action class. The idea of such an extension to videos could be  applied on top of various state-of-the-art object detectors. Here, we apply our method on top of SSD, as it has lower runtime than other detectors, which makes it suitable for large video datasets. In this section, %after briefly presenting SSD (Section~\ref{sub:ssd}), 
we first describe our fdafd4dfposed ACT-detector (Section~\ref{sub:sstd}), and then our full framework for video detection (Section~\ref{sub:video}). 
Finally, Section~\ref{sub:tubes} describes our method for constructing action tubes.

\begin{figure*}[t]
\centering
\includegraphics[width=\linewidth]{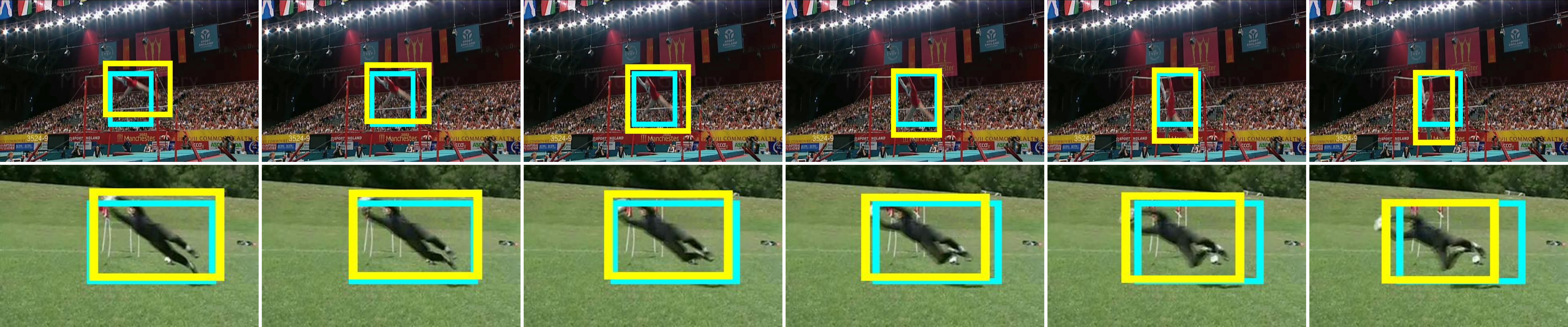}
\vspace{-4mm}
\caption{
Examples of regressed tubelets (yellow) from cuboids (cyan) in our ACT-detector. Note the accurate localization of the tubelet, despite the fact that the aspect ratio of the cuboid is changing over time.}
\label{fig:regression}
\vspace{-5mm}
\end{figure*}

%\vspace{-1mm}
\subsection{ACT-detector}
\label{sub:sstd}

In this paper, we claim that action localization benefits from predicting tubelets taking as input a sequence of frames instead of operating at the frame level. 
Indeed, the appearance and even the motion may be ambiguous for a single frame. 
Considering more frames for predicting the scores reduces this ambiguity (Figure~\ref{fig:splash}). 
Moreover, this allows to perform regression jointly over consecutive frames. % instead of doing it independently for each of them.
Our ACT-detector builds upon SSD, see Figure~\ref{fig:main} for an overview of the approach. 
In the following we review the SSD detector in details and then present our ACT-detector.

\paragraphV{SSD detector.~}The SSD detector (Single Shot MultiBox Detector)~\cite{liu16eccv} performs object detection by considering a
set of anchor boxes at different positions, scales and aspect ratios. 
Each of them is (a)~scored for each object class and for a background class, and (b)~regressed to better fit the object extent.  
SSD uses a fully convolutional architecture, without any object proposal step, enabling fast computation. Classification and regression are performed using different convolutional layers depending on the scale of the anchor
box. 
Note that the receptive field of a neuron used to predict the classification scores and the regression of a given anchor box remains significantly larger than the box. 

\paragraphV{ACT-detector.~}Given a sequence of $K$ frames, the ACT-detector computes convolutional features for each one. The weights of these convolutional features are shared among all input frames. 
We extend the anchor boxes of SSD to anchor cuboids by assuming that the spatial extent is fixed over time along the $K$ frames. 
We then stack the corresponding convolutional features from each of $K$ frames (Figure~\ref{fig:main}). 
The stacked features are the input of two convolutional layers, one for scoring action classes and one for regressing the anchor cuboids. 
For instance, when considering an anchor cuboid for which the prediction is based on the `red' feature maps of Figure~\ref{fig:main}, the classification and regression are performed with convolutional layers that take as input the `red' stacked feature maps from the $K$ frames. 
The classification layer outputs for each anchor cuboid $C+1$ scores: one per action class plus one for the background. This means that the tubelet classification is done based on the sequence of frames. 
The regression outputs $4 \times K$ coordinates (4 for each of the $K$ frames) for each anchor cuboid. Note that although all boxes in a tubelet are regressed jointly, they result in a different regression for each frame.

The initial anchor cuboids have a fixed spatial extent over time. In Section~\ref{sub:cuboid} we show experimentally that such anchor cuboids can handle moving actors for short sequences of frames. Note that the receptive field of the neurons used to score and regress an anchor cuboid is larger than its spatial extent. This allows us to base the prediction also on the context around the cuboid, \ie, with knowledge for actors that may move outside the cuboid. Moreover, the regression significantly deforms the cuboid shape. Even though anchor cuboids have fixed spatial extent, the tubelets obtained after regressing the $4 \times K$ coordinates do not. We display two examples in Figure~\ref{fig:regression} with the anchor cuboid (cyan boxes) and the resulting regressed tubelet (yellow boxes). Note how the regression outputs an accurate localization despite the change in aspect ratio of the action boxes across time.

\paragraphV{Training loss.} 
For training, we consider only sequences of frames in which all frames contain the ground-truth action. As we want to learn action tubes, all positive and negative training data come from sequences in which actions occur. 
We exclude sequences in which the action starts or ends. Let $\mathcal{A}$ be the set of anchor cuboids. We denote by $\mathcal{P}$ the set of anchor cuboids for which at least one ground-truth tubelet has an overlap over $0.5$, and by $\mathcal{N}$ the complementary set. Overlap between tubelets is measured by averaging the Intersection over Union (IoU) between boxes over $K$ frames. 
Each anchor cuboid from $\mathcal{P}$ is assigned to ground-truth boxes with IoU over $0.5$. More precisely, let $x_{ij}^y \in \{0,1\}$ be the binary variable
whose value is $1$ if and only if the anchor cuboid $a_i$ is assigned to the ground-truth tubelet $g_j$ of label $y$.
The training loss $\mathcal{L}$ is defined as:
%\vspace{-2mm}
\begin{small}
\vspace{0.6mm}
\begin{equation}
\mathcal{L} = \frac{1}{N} \big( \mathcal{L}_{\text{conf}} + \mathcal{L}_{\text{reg}} \big)~~,
\end{equation}
\vspace{-5.3mm}
\end{small}
\noindent with \begin{small}$N=\sum_{i,j,y} x_{ij}^y$ \end{small} the number of positive assignments and \begin{small} $\mathcal{L}_{\text{conf}}$ \end{small} (resp.\, \begin{small} $\mathcal{L}_{\text{reg}}$\end{small}) the confidence (resp. regression) loss as defined below.

The confidence loss is defined using a softmax loss. Let $\hat{c}_i^y$ be the predicted confidence score (after softmax) of an anchor $a_i$ for class $y$. The confidence loss is:

%\vspace{-2mm}
\begin{small}
\vspace{0mm}
\begin{equation}
%\vspace{-2mm}
\mathcal{L}_{\text{conf}} = - \sum_{i \in \mathcal{P}} x_{ij}^y \log \left( \hat{c}_i^y \right) - \sum_{i \in \mathcal{N}} \log \left( \hat{c}_i^0 \right)~~.
\end{equation}
\vspace{-1mm}
\end{small}

The regression loss is defined using a Smooth-L1 loss between the predicted regression and the ground-truth target. We regress an offset for the center $(x,y)$ of each box in the tubelet, as well as for the width $w$ and the height $h$. The regression loss is averaged over $K$ frames. 
More precisely, let $\hat{r}_i^{x_k}$ be the predicted regression for the $x$ coordinate of anchor $a_i$ at frame $f_k$ and let $g_j$ be the ground-truth target. The regression loss is defined as:

%\vspace{-2mm}
\begin{footnotesize}
\vspace{-4mm}
\begin{equation}
\begin{gathered}
\mathcal{L}_{\text{reg}} 
=\frac{1}{K} \sum_{i \in \mathcal{P}}  \sum_{c \in \{x,y,w,h\}} x_{ij}^y 
\sum_{k=1}^K  \text{SmoothL1} \left( \hat{r}_i^{c_k} - \mathfrak{g}_{ij}^{c_k} \right)~~,\\
\text{with} ~~~~ \mathfrak{g}_{ij}^{x_k} = \frac{g_j^{x_k} - a_i^{x_k} }{a_i^{w_k}}   
\qquad \mathfrak{g}_{ij}^{y_k} = \frac{g_j^{y_k} - a_i^{y_k} }{a_i^{h_k}}~, \\
~~~~~~~\mathfrak{g}_{ij}^{w_k} = \log \left( \frac{g_j^{w_k}}{a_i^{w_k}} \right) 
\qquad \mathfrak{g}_{ij}^{h_k} = \log \left( \frac{g_j^{h_k}}{a_i^{h_k}} \right)~~.
\end{gathered}
\end{equation}
\vspace{-2mm}
\end{footnotesize}

\begin{figure*}[t]
\centering
\includegraphics[width=0.95\linewidth]{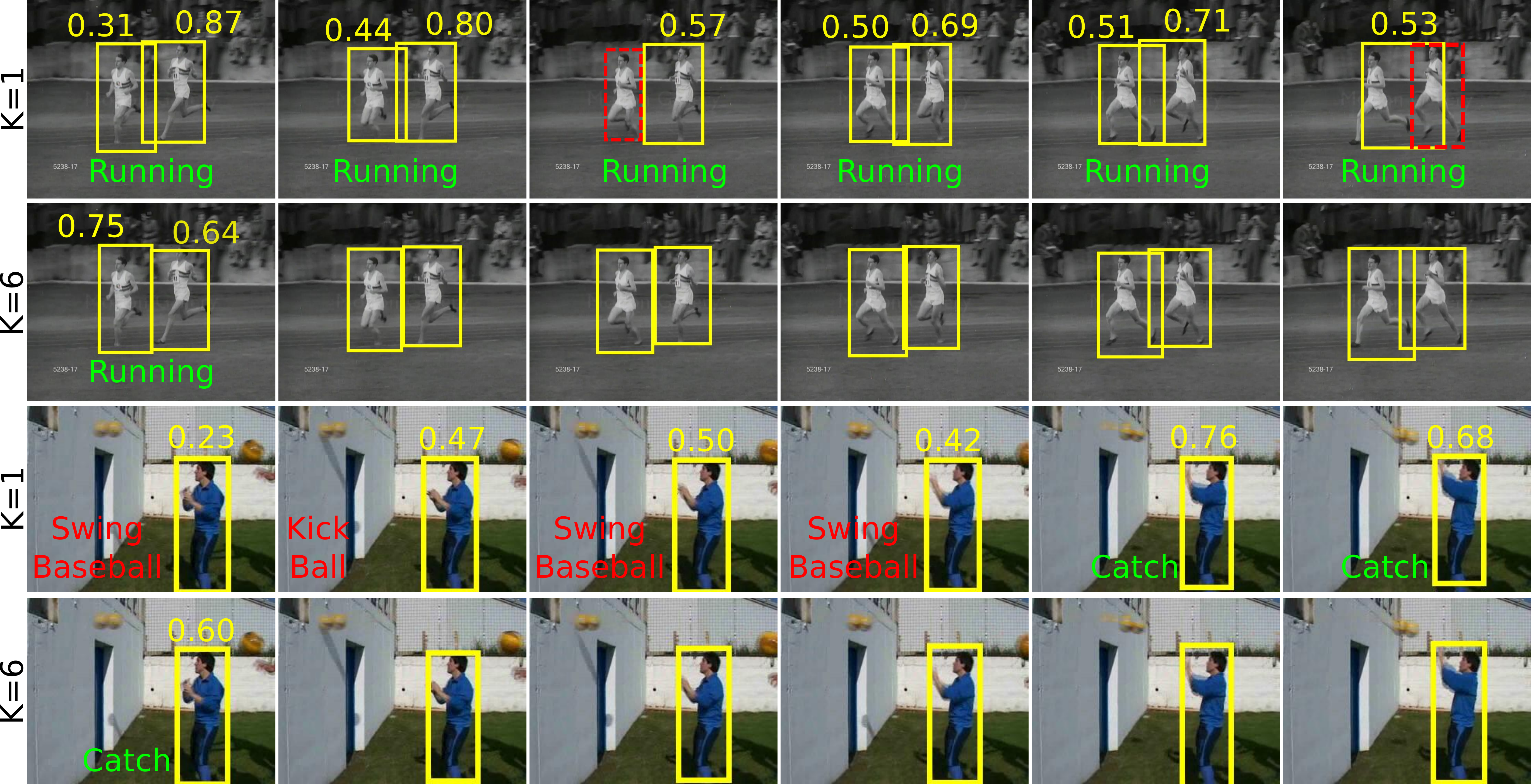} 
\vspace{-0.5mm}
\caption{
Examples when comparing per-frame ($K\!=\!1$) and tubelet detections ($K\!=\!6$). 
The yellow color represents the detections and their scores for the classes shown, the red color highlights errors due to missed detections (first row) or wrong labeling (third row) and the green color corresponds to correct labels. 
Our ACT-detector outputs one class label with one score per tubelet, we thus display it once.}
\vspace{-5.5mm}
\label{fig:tubelet}
\end{figure*}

%\vspace{-1mm}
\subsection{Two-stream ACT-detector}
\label{sub:video}

Following standard practice for action localization~\cite{Peng16eccv,Suman16bmvc,Weinzaepfel15iccv}, we use a two-stream detector. We train an appearance detector, for which the input is a sequence of $K$ consecutive RGB frames, and a motion detector, which takes as input the flow images~\cite{brox04eccv} obtained following~\cite{Gkioxari15cvpr}. 

Each stream outputs a set of regressed tubelets with confidence scores, originating from the same set of anchor cuboids. For combining the two streams at test time we compare two approaches: union fusion and late fusion. For the union fusion~\cite{singh16arxiv}, we consider the set union of the outputs from both streams: the tubelets from the RGB stream with their associated scores and the tubelets from the flow stream with their scores. For the late fusion \cite{feichtenhofer16cvpr}, we average the scores from both streams for each anchor cuboid, as the set of anchors is the same for both streams. We keep the regressed tubelet from the RGB stream, as appearance is more relevant for regressing boxes, in particular for actions with limited motion.  Our experiments show that late fusion outperforms the union fusion (Section~\ref{sub:modality}).  

%\vspace{-1mm}
\subsection{From action tubelets to spatio-temporal tubes}
\label{sub:tubes}

For constructing action tubes, we build upon the frame linking algorithm of~\cite{singh16arxiv}, as it is robust to missed detections and can generate tubes spanning different temporal extents of the video. 
%\ie, it already incorporates temporal detection. 
We extend their algorithm from frame linking to \textit{tubelet linking} and propose a \textit{temporal smoothing} to build action tubes from the linked tubelets. The method is online and proceeds by iteratively adding tubelets to a set of links while processing the frames. In the following, $t$ is a tubelet and $L$ a link, \ie, a sequence of tubelets.

\paragraphV{Input tubelets.} 
Given a video, we extract tubelets for each sequence of $K$ frames. This means that consecutive tubelets overlap by $K\!-\!1$ frames. The computation of overlapping tubelets can be performed at an extremely low cost as the weights of the convolutional features are shared. We compute the convolutional features for each frame only once. For each sequence of frames, only the last layers that predict scores and regressions, given the stacked convolutional features (Figure~\ref{fig:main}), remain to be computed. For linking, we keep only the $N\!=\!10$ highest scored tubelets for each class after non-maximum suppression (NMS) at a threshold $0.3$ in each sequence of frames.

\paragraphV{Overlap between a link and a tubelet.} 
Our linking algorithm relies on an overlap measure $\text{ov}(L,t)$ between a link $L$ and a tubelet $t$ that temporally overlaps with the end of the link. We define the overlap between $L$ and $t$ as the overlap between the last tubelet of the link $L$ and $t$. The overlap between two tubelets is defined as the average IoU between their boxes over overlapping frames.

\paragraphV{Initialization.} 
In the first frame, a new link is started for each of the $N$ tubelets. 
At a given frame, new links start from tubelets that are not associated to any existing link.

\paragraphV{Linking tubelets.} 
Given a new frame $f$, we extend one by one in descending order of scores each of the existing links with one of the $N$ tubelet candidates starting at this frame. The score of a link is defined as the average score of its tubelets. 
To extend a link $L$, we pick the tubelet candidate $t$ that meets the following criteria: (i)~is not already selected by another link, (ii)~has the highest score, and (iii)~verifies $\text{ov}(L,t)\!\geqslant\!\tau$, with $\tau$ a given threshold. In our experiments we use $\tau\!=\!0.2$.

\paragraphV{Termination.} 
Links stop when these criteria are not met for more than $K-1$ consecutive frames. 

\paragraphV{Temporal smoothing: from tubelet links to action tubes.} For each link $L$, we build an action tube, \ie, a sequence of bounding boxes. The score of a tube is set to the score of the link, \ie, the average score over the tubelets in the link. To set the bounding boxes, note that we have multiple box candidates per frame as the tubelets are overlapping. One can simply use the box of the highest scored tubelet. Instead,  we propose a temporal smoothing strategy. For each frame, we average the box coordinates of tubelets that pass through that frame. This allows us to build smooth tubes. 

\paragraphV{Temporal detection.~}The initialization and termination steps result in tubes spanning different temporal extents of the video. 
Each tube determines, thus, the start and end in time of the action it covers. No further processing is required for temporal localization.

%% file: experiments.tex
%\vspace{-1mm}
\section{Experimental results}
\label{sec:experiments}

\begin{figure*}[t]
\centering
\begin{tabular}{c@{}c@{}c@{}c@{}c@{}}
\includegraphics[width=0.235\linewidth]{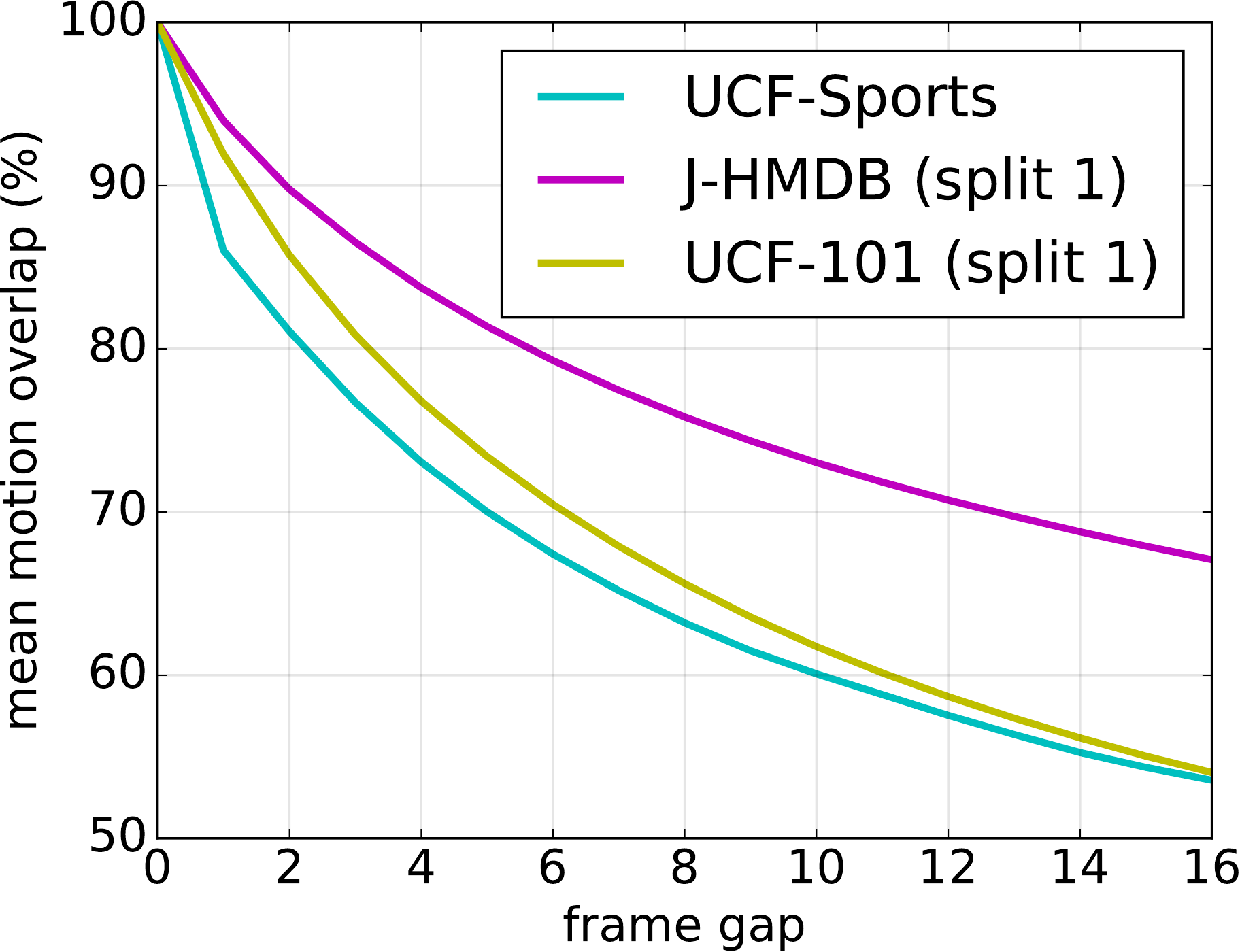}&
\textit{ ~~~~~} &
\includegraphics[width=0.235\linewidth]{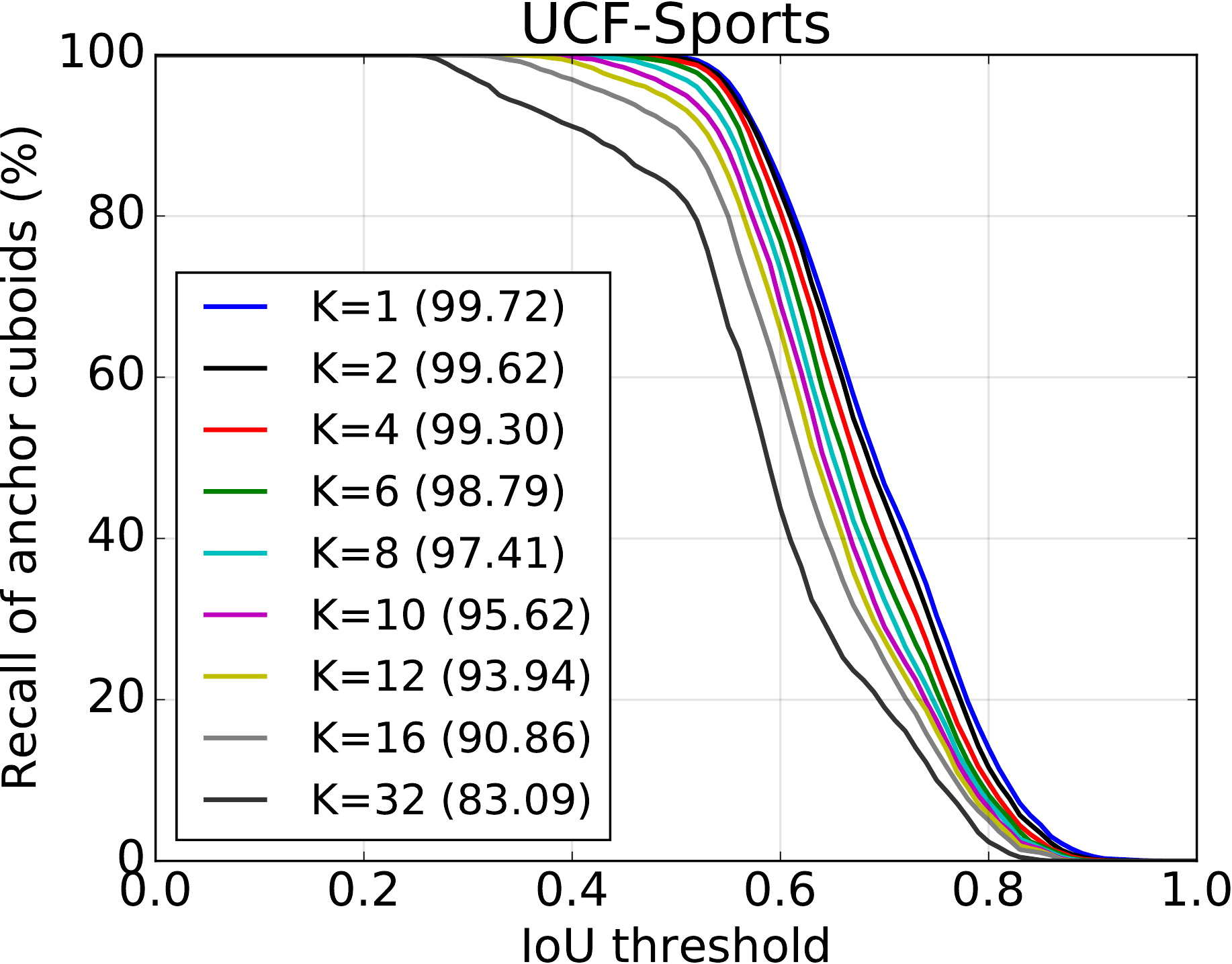} &
\includegraphics[width=0.235\linewidth]{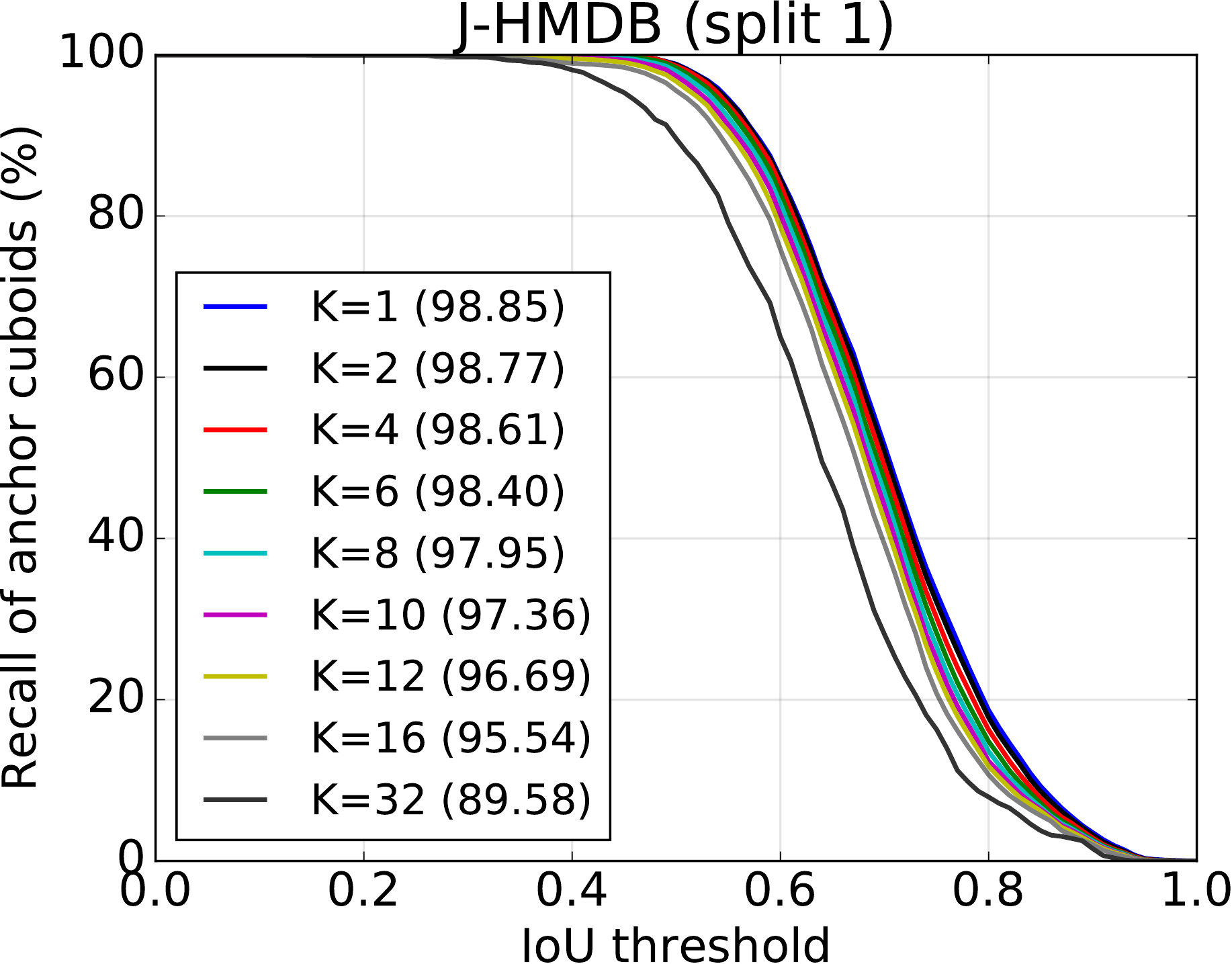} & 
\includegraphics[width=0.235\linewidth]{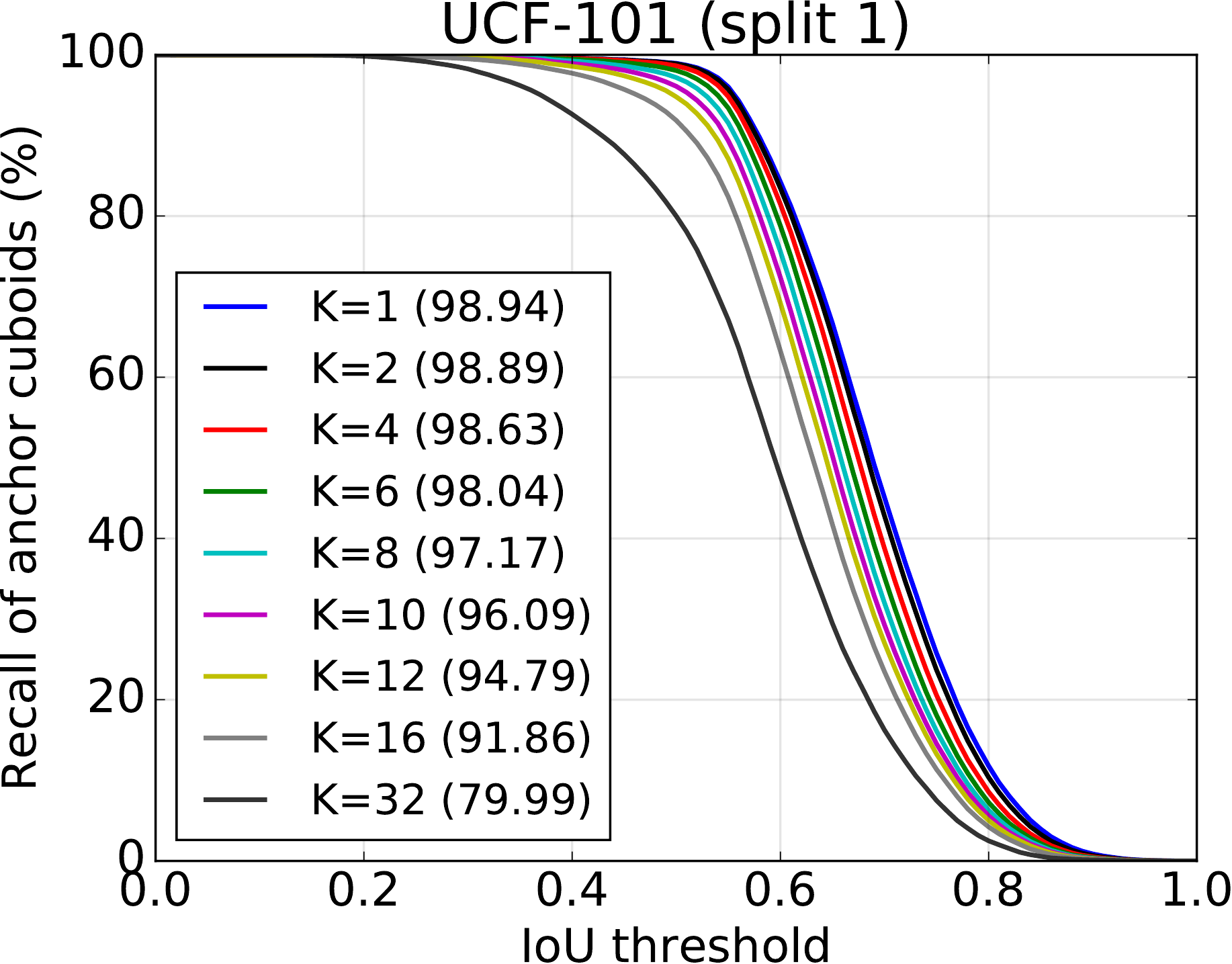} \\[-1mm]
(a) & & (b) & (c) & (d) 
\end{tabular}
\vspace{-0mm}
\caption{(a)~\textit{Motion overlap}: Mean motion overlap between a box in a ground-truth tube and its box $n$ frames later for varying $n$. (b-d)~Recall of the anchor cuboids for various IoU thresholds on the training set of three action localization datasets. The numbers in parenthesis indicate the recall at $\text{IoU}\!=\!0.5$. }
\label{fig:cuboids}
\vspace{-5mm}
\end{figure*} 

In this section we study the effectiveness of our ACT-detector.
After presenting the datasets used in our experiments (Section~\ref{sub:datasets}), we provide an analysis of our ACT-detector:  
we validate anchor cuboids (Section~\ref{sub:cuboid}), evaluate input modalities (RGB and flow) and their fusion (Section~\ref{sub:modality}),  
examine the impact of the length $K$ of the sequence of frames (Section~\ref{sub:snippet}), 
and present an error analysis (Section~\ref{sub:error}). 
We finally compare our method to the state of the art (Section~\ref{sub:comparison}).   

%\vspace{-1mm}
\subsection{Datasets, metrics and implementation details}
\label{sub:datasets}

\paragraphV{Datasets.~}The \textit{UCF-Sports} dataset~\cite{ucfsports} contains $150$ videos from $10$ sports classes such as \textit{diving} or \textit{running}. 
The videos are trimmed to the action. We use the train/test split of~\cite{lan11iccv}. 

The \textit{J-HMDB} dataset~\cite{jhmdb} contains $928$ videos with $21$ actions, including \textit{brush hair} and \textit{climb stairs}. The videos are trimmed to the action. 
We report results averaged on the three splits defined in~\cite{jhmdb}, unless stated otherwise.  

The \textit{UCF-101} dataset~\cite{ucf101} contains spatio-temporal annotations for $24$ sports classes in $3207$ videos. The videos are not trimmed. 
Following~\cite{Gkioxari15cvpr,Peng16eccv,Suman16bmvc,Weinzaepfel15iccv}, we report results for the first split only. 

\paragraphV{Metrics.~}We use metrics at both frame and video level. 
Frame-level metrics allow us to compare the quality of the detections independently of the linking strategy. 
Metrics at the video level are the same as the ones at the frame level, replacing the Intersection-over-Union (IoU) between boxes by a spatio-temporal overlap between tubes, \ie, an average across time of the per-frame IoU~\cite{Suman16bmvc,Weinzaepfel15iccv}. 
To measure our performance at the frame level, we take into account the boxes originating from all tubelets that pass through the frame with their individual scores and perform NMS. 
In all cases, we only keep the detections with a score above $0.01$. 

We report \textit{frame} and \textit{video mean Average Precision (mAP)}. A detection is correct if its IoU with a ground-truth box or tube is greater than $0.5$ and its action label is correctly predicted~\cite{pascal07:thomas}. For each class, we compute the average precision (AP) and report the average over all classes.

To evaluate the localization accuracy of the detections, we report \textit{MABO} (Mean Average Best Overlap)~\cite{uijlings13ijcv}. We compute the IoU between each ground-truth box (or tube) and our detections. For each ground truth box (or tube), we keep the overlap of the best overlapping detection (BO) and, for each class, we average over all boxes (or tubes) (ABO). The mean is computed over all classes (MABO).  

To evaluate the quality of the detections in terms of scoring, we also measure \textit{classification accuracy}. In each frame, assuming that the ground-truth localization is known, we compute class scores for each ground-truth box by averaging the scores from the detected boxes or tubelets (after regression) whose overlap with the ground-truth box of this frame is greater than $0.7$. We then assign the class having the highest score to each of these boxes and measure the ratio of boxes that are correctly classified.

%\vspace{-1mm}
\paragraphV{Implementation details.~}We use VGG~\cite{simonyan15iclr} with ImageNet pre-training for both appearance and motion streams~\cite{Peng16eccv,wang16eccv}. 
Our frames are resized to $300 \times 300$. 
We use the same hard negative mining strategy as SSD \cite{liu16eccv}, \ie, to avoid an unbalanced factor between positive and negative samples, only the hardest negatives up to a ratio of 3 negatives for 1 positive are kept in the loss. 
We perform data augmentation to the whole sequence of frames: photometric transformation, rescaling and cropping. 
Given the $K$ parallel streams, the gradient of the shared convolutional layers is the sum over the $K$ streams. 
We find that dividing the learning rate of the shared convolutional layers by $K$ helps convergence, as it prevents large gradients.

%\vspace{-1mm}
\subsection{Validation of anchor cuboids}
\label{sub:cuboid}

This section demonstrates that an anchor cuboid \textit{can} handle moving actions. We first measure how much the actors move in the training sets of the three action localization datasets by computing the \textit{mean motion overlap}. For each box in a ground-truth tube, we measure its \textit{motion overlap}: the overlap between this box and the ground-truth box $n$ frames later for varying $n$. For each class, we compute the average motion overlap over all frames and we report the mean over all classes in Figure~\ref{fig:cuboids}~(a). 
We observe that the motion overlap reduces as $n$ increases, especially for UCF-Sports and UCF-101 for which the motion overlap for a gap of $n\!=\!10$ frames is around $60\%$. This implies that there is still overlap between the ground-truth boxes that are separated by $n\!=\!10$ frames. 
It also means that in many cases, this overlap is below $50\%$ due to the motion of the actor.  

In practice, we want to know if we have positive training anchor cuboids. Positive cuboids are the ones that have an overlap of at least $50\%$ with a ground-truth tubelet; the overlap being the average IoU between boxes over the $K$ frames in the sequence. Such cuboids are required for training the classifier and the regressor. Thus, we consider all possible training sequences and compute for each class the recall of the anchor cuboids with respect to the ground-truth tubelets, \ie, the ratio of ground-truth tubelets for which at least one anchor cuboid has an overlap over $0.5$. 
%We perform this experiment on the three action localization datasets and 
We report the mean recall over the classes for varying IoU thresholds for the three datasets in Figure~\ref{fig:cuboids}~(b-d). For all datasets, the recall at IoU$=\!0.5$ remains $\geqslant 98\%$ up to $K\!=\!6$ and over $95\%$ for $K\!=\!10$. This confirms that cuboid-shaped anchors can be used in case of moving actors. When increasing $K$, for instance to $32$, the recall starts dropping significantly. 

Given that sequences of up to $K\!=\!10$ frames result in high recall of the anchor cuboids, we examine in Sections~\ref{sub:modality} and \ref{sub:snippet} the performance of our tubelet detector for sequences of length ranging between $K\!=\!1$ and $K\!=\!10$.

%\vspace{-1mm}
\subsection{Tubelet modality}
\label{sub:modality}

In this section, we examine the impact of the RGB and flow modalities and their fusion on the performance of our ACT-detector. For all datasets, we examine the frame-mAP when using (i)~only RGB data, (ii)~only flow data, (iii)~union of RGB $+$ flow data~\cite{Suman16bmvc}, and (iv)~late fusion of RGB $+$ flow data for varying sequence length from $1$ to $10$ frames. 

\begin{figure}[t]
\includegraphics[width=\linewidth]{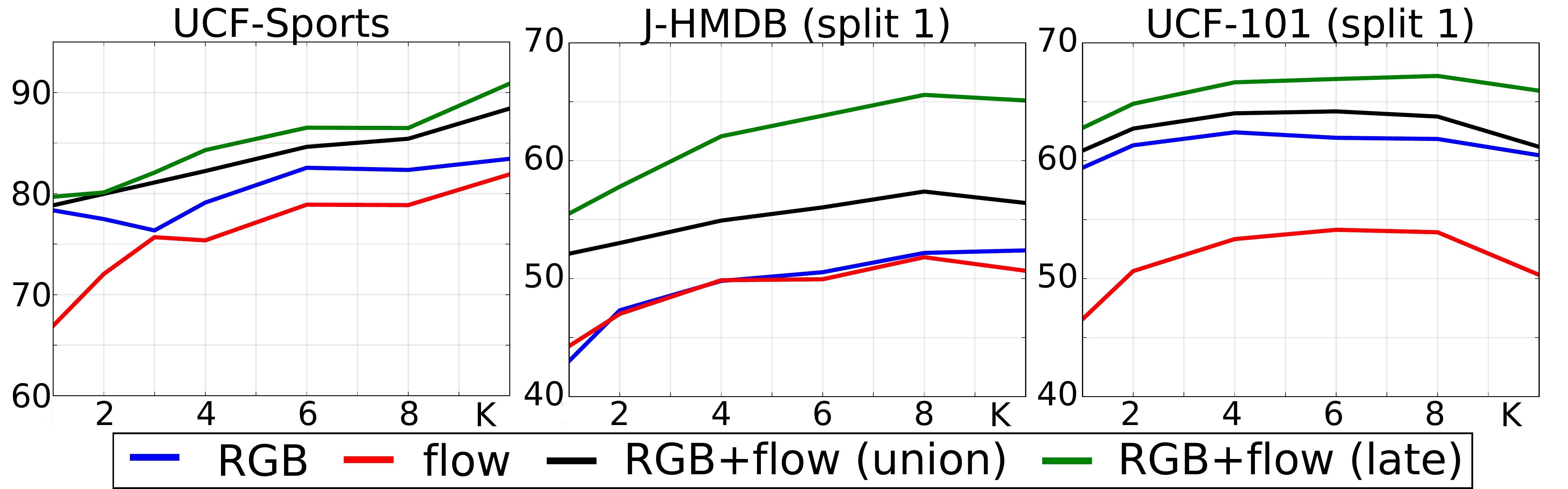}
\vspace{-4mm}
\caption{Frame-mAP of our ACT-detector on the three datasets when varying K for RGB data (blue line), flow (red line), union and late fusion of RGB $+$ flow data (black and green lines, resp.).}
\label{fig:frameAP}
\vspace{-5mm}
\end{figure}

For all datasets and for all $K$, the RGB stream (blue line) outperforms the flow stream (red line), showing that appearance information is on average a more distinctive cue than motion (Figure~\ref{fig:frameAP}).  
In all cases, using both modalities (green and black lines) improves the detection performance compared to using only one.  
We observe that late fusion of the scores (green line) performs consistently better than union fusion (black line), with a gain between $1\%$ and $4\%$ in terms of frame-mAP. This can be explained by the fact that union fusion considers a bigger set of detections without taking into account the similarity between appearance and motion detections. Instead, the late fusion re-scores every detection by taking into account both RGB and flow scores. Given that late fusion delivers the best performance, we use it in the remainder of this paper.

%\vspace{1mm}
\subsection{Tubelet length}
\label{sub:snippet}

In this section, we examine the impact of $K$. We consider $K\!=\!1$ as the baseline, and we report results for our method with $K\!=\!2,4,6,8,10$. We quantify the impact of $K$ by measuring (i)~the localization accuracy (MABO), (ii)~the classification accuracy, (iii)~the detection performance (frame-mAP), and (iv) the motion speed of actors. 

\begin{figure}
\centering
\includegraphics[width=\linewidth]{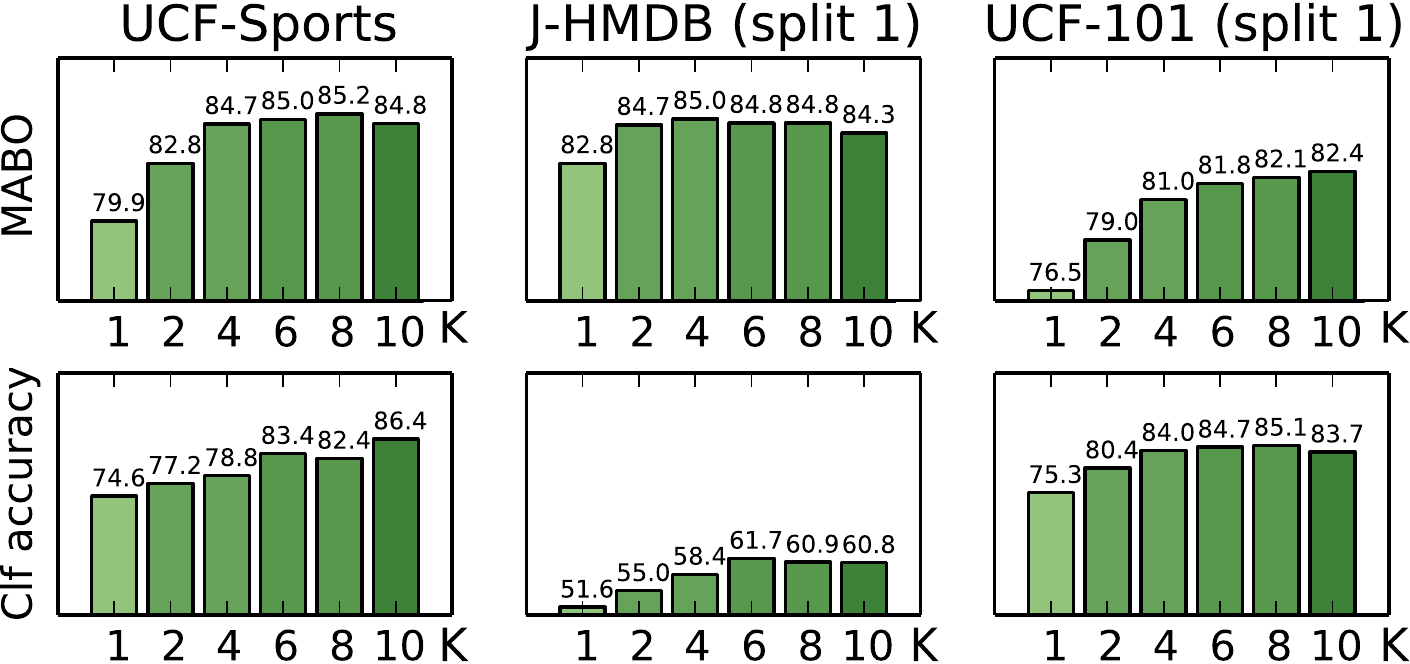} 
\vspace{-4mm}
\caption{MABO (top) and classification accuracy (bottom) of our ACT-detector on the three datasets when varying $K$.}
\label{fig:recallmabo}
\vspace{-4mm}
\end{figure}

\paragraphV{MABO. }  
MABO allows us to examine the localization accuracy of the per-frame detections when varying $K$. Results are reported in Figure~\ref{fig:recallmabo}~(top). For all three datasets we observe that using sequences of frames ($K\!>\!1$) leads to a significant improvement. In particular, MABO increases up to $K\!=\!4$, and then remains almost constant up to $K\!=\!8$ frames. For instance, MABO increases by $5\%$ on UCF-Sports, $2\%$ on J-HMDB and $5\%$ on UCF-101 when using $K\!=\!6$ instead of $K\!=\!1$. This clearly demonstrates that performing detection at the sequence level results in more accurate localization, see Figure~\ref{fig:regression} for examples. Overall, we observe that $K\!=\!6$ is one of the values for which MABO obtains excellent results for all datasets. 

\paragraphV{Classification accuracy. } 
We report classification accuracy on the three action localization datasets in Figure~\ref{fig:recallmabo}~(bottom). Using sequences of frames ($K\!>\!1$) improves the classification accuracy of the detections for all datasets. 
For UCF-Sports, the accuracy keeps increasing with $K$, while for J-HMDB it remains almost constant after $K\!=\!6$. 
For UCF-101, the accuracy increases when moving from $K\!=\!1$ to $K\!=\!4$ and after $K\!=\!8$ it starts decreasing. 
Overall, using up to $K\!=\!10$ frames improves performance over $K\!=\!1$. This shows that the tubelet scoring improves the classification accuracy of the detections. Again, $K\!=\!6$ is one of the values which results in excellent results for all datasets. 
\ifx
We report classification accuracy on the three action localization datasets in Figure~\ref{fig:recallmabo}~(bottom). Using sequences of frames ($K\!>\!1$) improves the classification accuracy of the detections for all datasets. For UCF-Sports, the accuracy keeps increasing with $K$, while for J-HMDB it remains almost constant after $K\!=\!6$. For UCF-101, we observe an increment when moving from $K\!=\!1$ to $K\!=\!4$ and then the accuracy starts decreasing. 
Overall, using up to $K\!=\!10$ frames improves performance over $K\!=\!1$. This shows that the tubelet scoring improves the classification accuracy of the detections. Again, $K\!=\!6$ is a value which results in excellent results for all datasets. 
\fi

\begin{table}[t]
\centering
\resizebox{0.99\linewidth}{!}{
\begin{tabular}{|c||c|c|c|c|c|c|c|c|c|c|}
\hline
\multirow{2}{*}{$K$} & \multicolumn{3}{c||}{UCF-Sports}  & \multicolumn{3}{c||}{J-HMDB (split 1)} &  \multicolumn{3}{c|}{UCF-101 (split 1)}  \\ 
\cline{2-10}
& slow & medium & \multicolumn{1}{c||}{fast} & slow & medium & \multicolumn{1}{c||}{fast} & slow & medium & fast \\
\hline \hline
$K\!=\!1$ & 79.5  & 84.0 & \multicolumn{1}{c||}{68.1} & 61.2 & 55.5 & \multicolumn{1}{c||}{49.0} & 69.6 & 73.5 & 67.3 \\
$K\!=\!6$ & 85.5  & 89.7 & \multicolumn{1}{c||}{76.8} & 69.8 & 66.9 & \multicolumn{1}{c||}{58.0} & 75.4 & 78.5 & 70.7 \\
\hline
\end{tabular}}
\vspace{1mm}
\caption{Frame-mAP for slow, medium and fast moving actors. %for $K\!=\!1$ and $K\!=\!6$. %We compute the motion of the actors using the mean motion overlap.
}
\label{table:largedisplacements}
\vspace{-5mm}
\end{table}

\paragraphV{Frame-mAP.~}Figure~\ref{fig:frameAP} shows the frame-mAP when training the ACT-detector with varying $K$. On all three datasets, we observe a gain up to $10\%$ when increasing the tubelet length up to $K\!=\!6$ or $8$ frames depending on the dataset, compared to the standard baseline of per-frame detection. This result highlights the benefit of performing detection at the sequence level. For J-HMDB and UCF-101, we also observe a performance drop for $K\!>\!8$, because 
%This can be explained by the fact that 
regressing from anchor cuboids is harder as (a)~the required transformation is larger when the actor moves, and (b)~there are more training parameters for less positive samples, given that the recall of anchor cuboids decreases (Section~\ref{sub:cuboid}).
The above results show that $K\!=\!6$ gives overall good results. 
We use this value in the following sections. 

Figure~\ref{fig:tubelet} shows some qualitative examples comparing the performance for $K\!=\!1$ and $K\!=\!6$.  
We observe that our tubelets lead to less missed detections and to more accurate localization compared to per-frame detection (first and second rows). 
Moreover, our ACT-detector reduces labeling mistakes when one frame is not enough to disambiguate between classes. 
For instance, in the last row we predict the correct label \textit{catch}, whereas in the third row there is a big variance in the labels (\textit{swing basketball, kick ball, catch}).

\paragraphV{Handling moving actors.}~To validate that our ACT-detector can handle moving actors, we measure frame-mAP with respect to the speed of the actor. 
We group actors into three categories (slow, medium, fast) with 1/3 of the data in each category. 
Speed is computed using the IoU of an actor with its instances in $\pm10$ neighboring frames. 
Table~\ref{table:largedisplacements} reports the frame-mAP at $\text{IoU}\!=\!0.5$ for the three categories. 
For all datasets there is a clear gain between $K\!=\!1$ and $K\!=\!6$ for all speeds. 
In particular, for actors with fast motion the gain is $+8\%$ for UCF-Sports, $+9\%$ for J-HMDB, and $+3\%$ for UCF-101. 
This confirms that our tubelets can successfully handle large displacements. 
A potential explanation is the fact that the receptive fields are significantly larger than the the spatial extent of the anchor cuboid.

\begin{figure}[t]
\centering
\includegraphics[width=\linewidth]{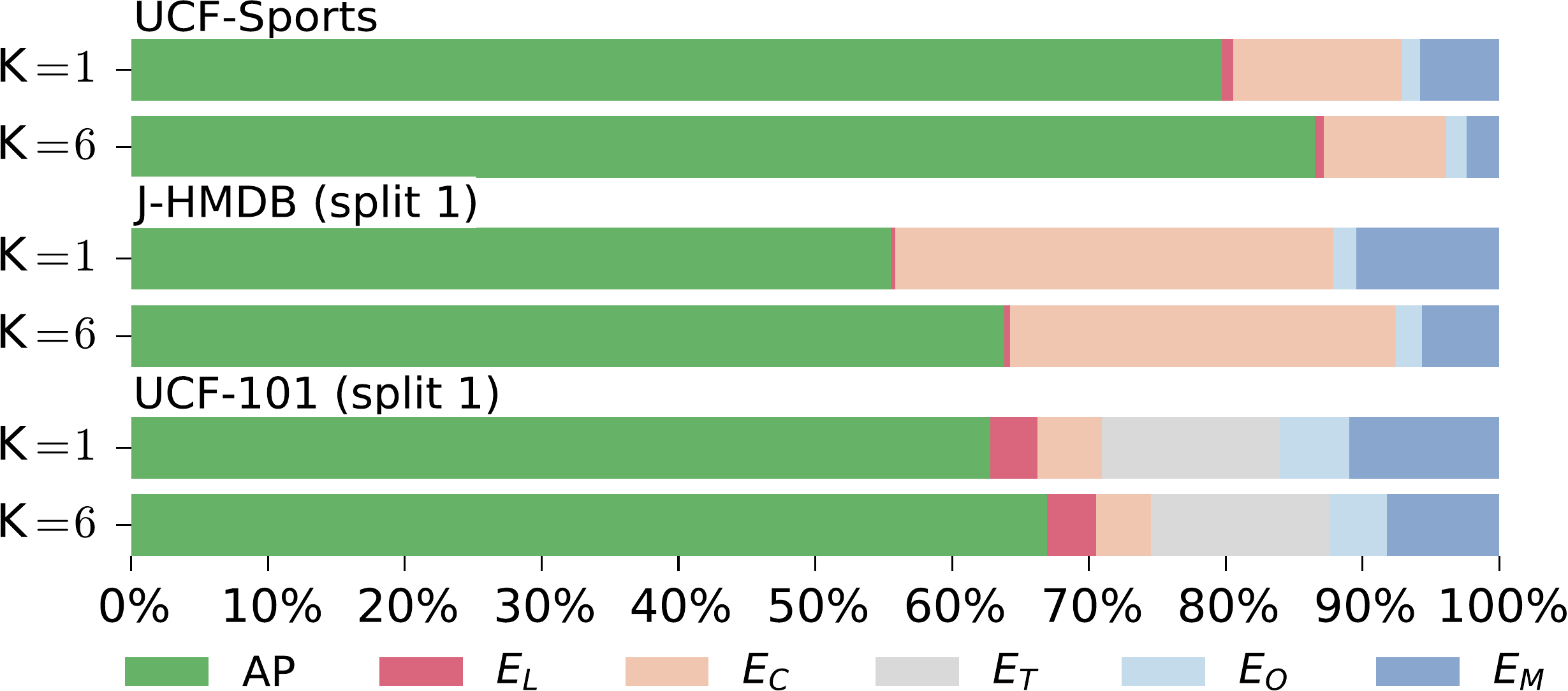} 
\vspace{-3.5mm}
\caption{Error analysis of our ACT-detector for $K\!=\!1$ and $K\!=\!6$ on three action localization datasets. We show frame-mAP and different sources of error, see Section~\ref{sub:error} for details.}
\vspace{-4.5mm}
\label{fig:error}
\end{figure}

%\vspace{-1mm}
\subsection{Error breakdown analysis} 
\label{sub:error}

In this section, we examine the cause of errors in frame-mAP to better understand the reasons why our tubelets improve detection performance. More precisely, we consider five mutually exclusive factors and analyze which percentage of the mAP is lost due to each of them:
\begin{enumerate}[noitemsep,nolistsep]
\item localization error $E_{L}$: the detection is in a frame containing the correct class, but the localization is wrong, \ie, $IoU<0.5$ with the ground-truth box.
\item classification error $E_{C}$: the detection has $IoU \geqslant 0.5$ with the ground-truth box of another action class. 
\item time error $E_{T}$: the detection is in an untrimmed video for the correct class, but the temporal extent of the action does not cover this frame. 
\item other errors $E_O$: the detection appears in a frame without the class, and has $IoU\!<\!0.5$ with ground-truth boxes of any other class.
\item missed detections $E_M$: we do not have a detection for a ground-truth box.  
\end{enumerate}

The first four factors are categories of false positive detections, while $E_M$ refers to the ones we did not detect at all. For the first four factors, we follow the frame-mAP computation and measure the area under the curve when plotting the percentage of each category at all recall values. The missed detections ($E_M$) factor is computed by measuring the percentage of missed detections, \ie, ratio of ground-truth boxes for which there are no correct detections.

Figure~\ref{fig:error} shows the percentage that each of these factors contributes to errors in the mAP for $K\!=\!1$ and $K\!=\!6$ with late fusion of RGB and flow as input modalities. For all datasets, we observe that when going from $K\!=\!1$ to $K\!=\!6$ there is almost no change in $E_{L}$ or in $E_O$. In particular, for UCF-Sports and J-HMDB their values are extremely small even for $K\!=\!1$. We also observe a significant decrease of $E_C$ between $K\!=\!1$ and $K\!=\!6$, in particular on the UCF-Sports and J-HMDB datasets. This highlights that including more frames facilitates the action classification task (Figure~\ref{fig:splash}). This drop is lower on the UCF-101 dataset. This can be explained by the fact that most errors in this dataset come from false detections outside the temporal extent of the actions ($E_T$). Note that $E_T\!=\!0$ on UCF-Sports and J-HMDB, as these datasets are trimmed. For all datasets, a big gain comes from the missed detections $E_M$: for $K\!=\!6$ the percentage of missed detections drops significantly compared to $K\!=\!1$. For instance, on J-HMDB the percentage of missed detections is reduced by a factor of 2. This clearly shows the ability of our proposed ACT-detector not only to better classify and localize ($E_{C}$ and MABO) actions but also to detect actions missed by the single-frame detector (see Figure~\ref{fig:tubelet}).
       
\begin{table}[t]
\centering
\resizebox{0.98\linewidth}{!}{
\begin{tabular}{|c|l||c|c|c|c|}
\hline
detector & method & UCF-Sports  & J-HMDB &  UCF-101 \\ 
\hline \hline
actionness & ~\cite{wang16cvpr} & - & 39.9 & - \\
\hline
\multirow{2}{*}{R-CNN} & ~\cite{Gkioxari15cvpr} & 68.1 & 36.2 & -\\ 
 & ~\cite{Weinzaepfel15iccv} &71.9 &45.8&35.8 \\ 
\cline{1-5}
Faster  & ~\cite{Peng16eccv} w/o MR & 82.3 & 56.9 & 64.8\\ 
R-CNN &  ~\cite{Peng16eccv} with MR & 84.5 & 58.5 & 65.7 \\ 
\cline{1-5} 
SSD & ~\textbf{ours} & \textbf{87.7} &\textbf{65.7} & \textbf{67.1}\\ 	
\hline
\end{tabular}}
\vspace{0.3mm}
\caption{Comparison of frame-mAP to the state of the art. For~\cite{Peng16eccv}, we report the results with and without their multi-region (+MR).% approach. 
%For J-HMDB we report results averaged over all splits and for UCF-101 we report results on the first split. 
}
\label{table:sotaframeAP}
\vspace{-6mm}
\end{table}       

\begin{table*}[!hbt]%[!htbp]%[t]
\centering
\resizebox{0.93\linewidth}{!}{
\begin{tabular}{|c|l||c|c|c|c||c|c|c|c||c|c|c|c|}
\hline
\multirow{2}{*}{detector} & \multirow{2}{*}{method} & \multicolumn{4}{|c||}{UCF-Sports}  & \multicolumn{4}{|c||}{J-HMDB (all splits)} &  \multicolumn{4}{|c|}{UCF-101 (split 1)}  \\ 
\cline{3-14} 
&  & 0.2 & 0.5 & 0.75 & 0.5:0.95 & 0.2 & 0.5 & 0.75 & 0.5:0.95 & 0.2 & 0.5 & 0.75 & 0.5:0.95\\ \hline \hline
actionness & ~\cite{wang16cvpr} &  - & - & - & - &  - & 56.4 & -& - & - & - & - & - \\
\hline
 \multirow{2}{*}{R-CNN} & ~\cite{Gkioxari15cvpr} & - & 75.8 & - & - & - & 53.3 & - & - &  - & - & - & -\\ 
 & ~\cite{Weinzaepfel15iccv} & - & 90.5 & - & - & 63.1 & 60.7 & - & - &  51.7 & - & - & -\\ 
\cline{1-14}

\multirow{3}{*}{Faster R-CNN}  & ~\cite{Peng16eccv} w/o MR & \textbf{94.8} & \textbf{94.8} & 47.3 & 51.0 & 71.1 & 70.6 & 48.2 & 42.2 &  71.8 & 35.9 & 1.6 & 8.8 \\ 
& ~\cite{Peng16eccv} with MR & \textbf{94.8} & 94.7 & - & - & \textbf{74.3} & \textbf{73.1} & - & - & 72.9 &-& - &- \\ 
 & ~\cite{Suman16bmvc} & - & - & - & - & 72.6 & 71.5 & 43.3 & 40.0 &  66.7 & 35.9 & ~7.9 & 14.4\\ 
 \cline{1-14}
\multirow{2}{*}{SSD}
 & ~\cite{singh16arxiv}  & - & - & - & - & 73.8 & 72.0 & 44.5 & 41.6 &  73.5 & 46.3& 15.0 & 20.4 \\  
 & ~\textbf{ours} & 92.7 & 92.7 & \textbf{78.4} & \textbf{58.8}
 & \textbf{74.2} & \textbf{73.7} & \textbf{52.1} & \textbf{44.8} &  
 \textbf{77.2} & \textbf{51.4} & \textbf{22.7} & \textbf{25.0} \\
\hline
\end{tabular}}
\vspace{0.3mm}
\caption{Comparison of video-mAP to the state of the art at various detection thresholds. The columns 0.5:0.95 correspond to the average video-mAP for thresholds with step $0.05$ in this range. For~\cite{Peng16eccv}, we report the results with and without their multi-region (+MR) approach. 
%For J-HMDB we report results averaged over all splits, and for UCF-101 we report results on the first split. 
}
\label{table:sotavideoAP}
\vspace{-5mm}
\end{table*}

\subsection{Comparison to the state of the art}
\label{sub:comparison}

We compare our ACT-detector to the state of the art. Note that our results reported in this section are obtained by stacking $5$ consecutive flow images~\cite{Peng16eccv,Simonyan2014nips} as input to the motion stream, instead of just $1$ for each of the $K\!=\!6$ input frames. This variant brings about $+1\%$ frame-mAP. 

\vspace{-0.2mm}
\paragraphV{Frame-mAP.~}We report frame-mAP on the three datasets in Table~\ref{table:sotaframeAP}. We compare our performance with late fusion of RGB$+$5flows when $K\!=\!6$ to~\cite{Gkioxari15cvpr,Weinzaepfel15iccv}, that use a two-stream R-CNN, and to~\cite{wang16cvpr}, which is based on actionness. We also compare to Peng and Schmid~\cite{Peng16eccv} that build upon a two-stream Faster R-CNN with multiscale training and testing. We report results of~\cite{Peng16eccv} with and without their multi-region approach. The latter case can be seen as the baseline Faster R-CNN with multiscale training and testing for $K\!=\!1$. Our ACT-detector (\ie, with $K\!=\!6$) brings a clear gain in frame-mAP, outperforming the state of the art on UCF-Sports, J-HMDB and UCF-101.
We also observe that overall the performance of the baseline SSD ($K\!=\!1$) is somewhat lower (by around $3$ to $5\%$) than Faster R-CNN used by the state of the art~\cite{Peng16eccv}, see Figure~\ref{fig:frameAP}. SSD, however, is much faster than Faster R-CNN, and therefore more suitable for large video datasets.

\vspace{-0.2mm}
\paragraphV{Video-mAP.~}Table~\ref{table:sotavideoAP} reports the video-mAP results for our method and the state of the art at various IoU thresholds ($0.2$, $0.5$, and $0.75$). We also report results with the protocol 0.5:0.95~\cite{coco}, which averages over multiple IoU thresholds, \ie, over $10$ IoU thresholds between $0.5$ and $0.95$ with a step of $0.05$. 
At rather low IoU thresholds ($0.2$, $0.5$) on UCF-Sports and J-HMDB the performance of our ACT-detector is comparable to the state-of-the-art methods that rely on Faster R-CNN \cite{Peng16eccv,Suman16bmvc} or on SSD ~\cite{singh16arxiv}. 
At higher overlap thresholds we significantly outperform them. 
For instance on UCF-Sports and J-HMDB at $\text{IoU}\!=\!0.75$ we outperform \cite{Peng16eccv} by $31\%$ and $4\%$. 
In particular, our performance drops slower than the state of the art as the IoU threshold increases.  This highlights the high localization accuracy of our tubelets and, therefore of our tubes. On UCF-101, we significantly outperform the state of the art at all overlap thresholds, with a larger gap at high thresholds. 
For instance, we outperform \cite{singh16arxiv} by $5\%$ at $\text{IoU}\!=\!0.5$, and by $7.5\%$ at $\text{IoU}\!=\!0.75$. 
To validate our tubelet linking strategy (Section~\ref{sub:tubes}), we experiment with an approach that transforms tubelets into individual boxes and links them with~\cite{Peng16eccv,Suman16bmvc}. 
We observe a consistent gain of $1\%$ on all datasets. 
As a summary, our ACT-detector improves over the state of the art, especially at high thresholds.

\vspace{-0.2mm}
\paragraphV{Runtime.~}We compare our runtime using two streams (appearance and flow) to the frame-based SSD approach of Singh~\etal~\cite{singh16arxiv} and to frame-based Faster R-CNN approaches~\cite{Peng16eccv,Suman16bmvc}. 
We report runtime on a single GPU without flow computation. 
Faster R-CNN based approaches~\cite{Peng16eccv,Suman16bmvc} run at 4fps and the SSD-based method~\cite{singh16arxiv} at 25-30fps.  
Our ACT-detector also runs at 25-30fps ($K\!=\!6$). Computing tubelets has a low overhead, since the convolutional features are computed once per frame due to the parallel architecture with shared weights. 
The post-processing is extremely fast ($\mathtt{\sim}$300fps) for all methods. 
%This highlights the clear benefit of our ACT-detector: running in real time and achieving higher accuracy than the state of the art. 

%% file: conclusions.tex
%\vspace{-1mm}
\section{Conclusions}
\label{sec:conclusions}

We introduced the ACT-detector, a tubelet detector that leverages the temporal continuity of video frames. 
It takes as input a sequence of frames and outputs \textit{tubelets} instead of operating on single frames, as is the case with previous state-of-the-art methods~\cite{Peng16eccv,Suman16bmvc,singh16arxiv}. 
Our method builds upon SSD and introduces anchor cuboids that are scored and regressed over sequences of frames. 
An extensive experimental analysis shows the benefits of our ACT-detector for both classification and localization. 
It achieves state-of-the-art results, in particular for high overlap thresholds.